# DIFFUSION AUTOENCODER FOR UNSUPERVISED ARTIFACT RESTORATION IN HANDHELD FUNDUS IMAGES

*Mathumetha Palani*[1]    *Kavya Puthumana*[1]    *Ayantika Das*[2]    *Ganapathy Krishnamurthi*[2]

[1]Nuvo AI, Chennai, India
[2]Indian Institute of Technology Madras, India

**ABSTRACT**

The advent of handheld fundus imaging devices has made ophthalmologic diagnosis and disease screening more accessible, efficient, and cost-effective. However, images captured from these setups often suffer from artifacts such as flash reflections, exposure variations, and motion-induced blur, which degrade image quality and hinder downstream analysis. While generative models have been effective in image restoration, most depend on paired supervision or predefined artifact structures, making them less adaptable to unstructured degradations commonly observed in handheld fundus images. To address this, we propose an unsupervised diffusion autoencoder that integrates a context encoder with the denoising process to learn semantically meaningful representations for artifact restoration. The model is trained only on high-quality table-top fundus images and infers to restore artifact-affected handheld acquisitions. We validate the restorations through quantitative and qualitative evaluations, and have shown that diagnostic accuracy increases to $81.17\%$ on an unseen dataset and multiple artifact conditions.

## 1. INTRODUCTION

Ophthalmologic analysis relies significantly on fundus imaging as a primary screening tool for various diagnostic purposes [1]. The continuous advancement in identifying fundus image–based biomarkers [2] for multiple diseases, such as diabetic retinopathy, has established fundus screening as a widely adopted and crucial diagnostic approach. These advancements have progressively transitioned fundus imaging from traditional table-top acquisition systems to compact handheld and mobile-based devices, enabling portable, cost-effective, and point-of-care retinal screening [3]. Typically, these handheld systems integrate a smartphone camera with a specialized optical attachment for direct fundus capture. Although these systems are more accessible and portable, the acquired images often suffer from artifacts such as over- or under-exposure, poor stabilization, and flash-induced reflections, which degrade image quality and hinder reliable downstream analysis.

The integration of AI-based modeling approaches for analyzing fundus images from table-top devices has significantly improved the efficiency of automated retinal assessment pipelines. The high-quality images from these systems enable AI models to effectively learn image distributions and perform disease detection tasks with high accuracy. However, extending similar AI strategies to handheld device images remains challenging, as artifacts and illumination inconsistencies often consider these images out-of-distribution for conventional assessment models [4]. *How can we leverage high-quality fundus images to restore artifact-degraded handheld acquisitions for reliable downstream analysis?*

The adoption of generative modeling approaches has been shown to be effective for various artifact restoration tasks. The Generative Adversarial Networks (GANs) learn to generate images while restoring missing pixels, learning from neighborhood contexts [5], [6]. While many of these models use predefined or free-form masks to explicitly inpaint missing regions, they typically perform better when the masks are such that the surrounding context is structurally intact.

Denoising diffusion models have demonstrated strong capability in modeling image distributions [7] and performing unsupervised artifact inpainting during inference [8]. This formulation alleviates dependence on predefined artifact structures, allowing training in an unconstrained manner. However, conventional diffusion processes in an unconstrained setup might not be able to preserve contextual information such as vascular structures partially occluded by artifacts. To address this, *we propose an unsupervised diffusion autoencoder with a context encoder integrating semantically meaningful information for artifact restoration in fundus images, leveraging high-quality table-top acquired images*. Our contributions are:

- We introduce a **diffusion auto-encoding** framework that integrates a context encoder with the denoising process to learn semantically meaningful representations for restoring artifacts in handheld fundus images.
- We further devise an **unsupervised** mechanism that learns only to generate high-quality table-top fundus images and infers to inpaint and restore artifact-affected fundus images acquired from handheld devices.
- We validate our approach through quantification on arti-

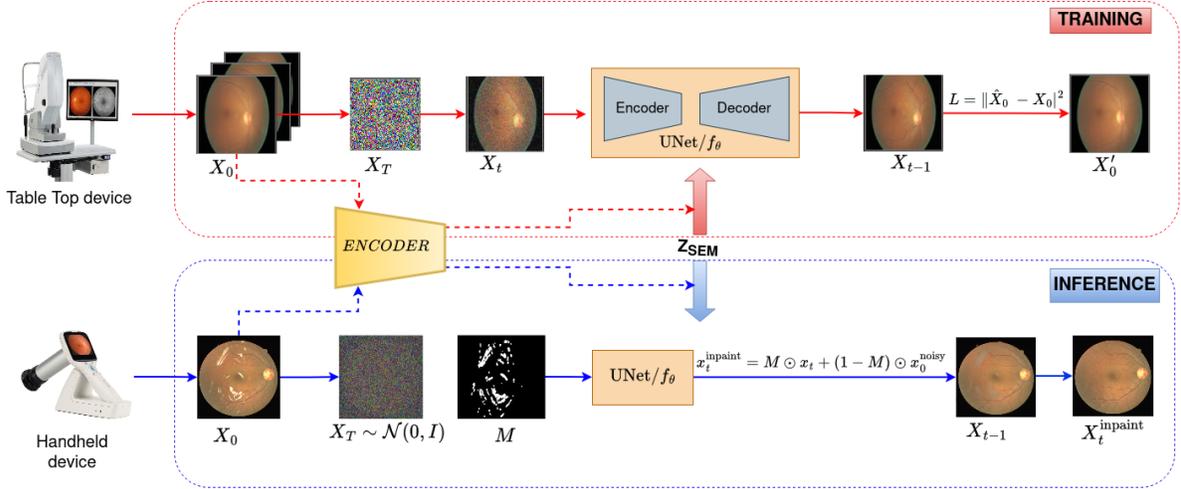

**Fig. 1**. Top to bottom: (i) The training pipeline learning to auto-encode only with table-top high-quality fundus images and (ii) the inference pipeline performing restoration through inpainting with fundus images from handheld devices.

fact datasets completely **unseen** during training and further perform **downstream** tasks like vessel segmentation and disease grading to evaluate our restorations.

**Related Work**: CNN-based models such as ShiftNet [9] and MISF [10] rely on encoder–decoder frameworks to reconstruct missing regions but struggle to retain fine retinal patterns. GAN-based methods achieved improved realism, with EdgeConnect [5] employing a two-stage edge-guided inpainting pipeline. DeepFillV2 [6] and CR-Fill [11] introduced gated convolutions and attention for context-aware filling by optimizing using a contextual reconstruction loss and a Patch-GAN discriminator [12]. Diffusion-based fundus restoration models, such as ClarityNet [1], utilize paired masks and clean images to train the denoising process. This might lack contextual awareness when the size of the masks is larger.

## 2. METHODOLOGY

Our proposed method employs a Diffusion Autoencoder (DiffAE) framework that integrates an encoder with the denoising process (UNet) to learn image representations through an auto-encoding objective on high-quality table-top fundus images, as detailed in Subsection 2.1. The trained DiffAE is then utilized for artifact inpainting in handheld fundus images. This inference-based inpainting strategy incorporates artifact masks at each denoising timestep to guide restoration [8], as discussed in Subsection 2.2. The architectural diagram of the training and inference process is given in Figure 1.

### 2.1. Diffusion Auto-encoder (DiffAE)

The DiffAE framework incorporates a learnable context encoder, $Enc_\phi$, with the denoising UNet, $f_\theta$. The encoder maps a clean fundus image $x_0$ into a compact latent vector $z = Enc_\phi(x_0)$, where $z \in \mathbb{R}^{512}$ [13]. This latent representation captures global retinal structures, serving as a semantic prior for the reverse diffusion process.

As in a conventional diffusion process, forward diffusion adds Gaussian noise to a clean image $x_0$ over $T$ time steps, yielding noisy samples $\{x_t\}_{t=1}^{T}$. The conditional distribution is $q(x_t \mid x_0) = \mathcal{N}(x_t; \sqrt{\bar{\alpha}_t} x_0, (1 - \bar{\alpha}_t)I)$, where $\bar{\alpha}_t$ represents the noise coefficients controlling variance schedule [7].

The reverse process reconstructs $x_0$ by learning the distribution, $p_\theta(x_{t-1} \mid x_t, z) = \mathcal{N}(x_{t-1}; f_\theta(x_t, t, z), \sigma_t^2 I)$, where $f_\theta$ is the denoising UNet conditioned on $x_t$ and latent $z$, while $\sigma_t^2$ determined by the predefined noise scheduler.

The model predicts $\hat{x}_0 = f_\theta(x_t, t, z)$ and minimizes the MSE loss $L_{\text{MSE}} = \|\hat{x}_0 - x_0\|^2$, ensuring accurate reconstruction and semantically rich latent embeddings [14]. The DiffAE model, conditioned on image representations $z$, learns to generate high-quality retinal structures that can effectively fill artifact masks in a contextual manner.

### 2.2. Artifact Restoration with DiffAE

For artifact restoration, inference is performed on the trained DiffAE by incorporating artifact masks $M$ as a conditioning in the reverse denoising process [8], guided by the contextual representation $z$ from the encoder, formulated as, $p_\theta(x_{t-1} \mid x_t, z, M) = \mathcal{N}(f_\theta(x_t, t, z, M), \sigma_t^2 I)$. The mask conditioning is enabled at each timestep $t$ through the following equation:

$$x_t^{\text{inpaint}} = M \odot x_t + (1 - M) \odot x_0^{\text{noisy}}, \quad (1)$$

where $x_0^{\text{noisy}}$ is the artifact image incorporated with noise.

*Artifact Mask*: Artifact masks ($M$) were created from handheld fundus images through identification, processing, and binarization of artifact regions. Restorations were performed on both real and simulated artifacts. For simulation, artifacts extracted from handheld images were separated into texture components, normalized, and blended with clean tabletop images while maintaining anatomical consistency, such as optic disc alignment.

Table 1. Quantification of our restoration output along with the comparative baselines for the *Test* and *Synthetic Set*.

| Model | PSNR(↑) (dB) | SSIM(↑) | Vessel Segmentation (Dice(↑)) | Quality Assessment (%) | | | Classification (Accuracy %) |
|---|---|---|---|---|---|---|---|
| | | | | Good(↑) | Usable(↑) | Reject(↓) | |
| EdgeConnect [5] | 36.04 ± 3.06 | 0.94 ± 0.03 | 0.706 | 14.7 | 26.3 | 59.0 | 80.04 |
| MISF [10] | 26.99 ± 1.02 | 0.91 ± 0.01 | 0.685 | 12.6 | 33.7 | 53.7 | 77.09 |
| ShiftNet [9] | 17.58 ± 2.67 | 0.67 ± 0.05 | 0.576 | 24.2 | 39.0 | 36.8 | 80.02 |
| ClarityNet [1] | 31.98 ± 2.54 | 0.93 ± 0.01 | 0.600 | 31.6 | 33.7 | 34.7 | 76.94 |
| DeepFillV2 [6] | 37.53 ± 2.20 | **0.97 ± 0.01** | 0.874 | 31.6 | 47.4 | 16.8 | 80.50 |
| **Ours** | **38.64 ± 0.35** | **0.97 ± 0.01** | **0.891** | **35.8** | **52.6** | **15.8** | **81.17** |

## 3. EXPERIMENTAL SETUP

**Dataset Details And Evaluation Metrics**: *Train Set*: We have utilized healthy fundus images from the EyePACS [15] dataset to train our model. This dataset comprises of $12,098$ high-quality healthy images acquired in a table-top setup. *Test Set*: We have utilized mobile fundus image dataset, mBRSET [16] for inference. From the dataset, 197 images containing artifacts like over-, under-exposed, and flash-related artifacts were identified and verified by an ophthalmologist. The restored images from this set were evaluated with a quality assessment score and the accuracy of a downstream diabetic retinopathy (DR) classification task. *Synthetic Set*: Further, we have synthesized 40 artifact-clean pairs on the the healthy images from DRIVE [17] dataset. Flash artifacts, particularly reflections near the optic disc and macular regions, were synthetically placed into the DRIVE set by extracting from the mobile dataset. The restored images were evaluated with image quality metrics PSNR and SSIM, and with the Dice score of the segmented vessels[1].

**Implementation Details and Baselines**: The proposed framework was implemented in PyTorch and trained on an NVIDIA H100 GPU with 80 GB memory. The Adam optimizer was employed with an initial learning rate of $1 \times 10^{-4}$, decayed linearly over 150 epochs. Training was conducted on images of size $512 \times 512$ with a batch size of 4. **Baselines**: We compare with MISF [10], EdgeConnect [5], ShiftNet [9], DeepFillV2 [6], and ClarityNet [1], covering multi-level CNN, edge-aware GAN, shift-layer GAN, gated-convolution GAN, and diffusion-based inpainting approaches. All baselines utilize official implementations trained on artifact-free, table-top images (*Train Set*) with random, free-form masks, except for the diffusion model, which was trained for image generation and evaluated using the inference-time inpainting strategy to align with our setup.

## 4. RESULTS AND DISCUSSION

### 4.1. Quantitative Analysis

In order to quantify the restoration quality, we evaluate (i) PSNR, SSIM, and Vessel segmentation score on *Synthetic Set* and (ii) Quality assessment score and DR classification accuracy on *Test Set* and report in Table 1.

*Image Quality Assessment on Synthetic Set*: The first two columns of Table 1 indicate the PSNR and SSIM on the *Synthetic Set* (with paired ground). Our model achieves a 1.11 dB higher PSNR than the best-performing baseline, DeepFillV2, with comparable SSIM. Among other methods, the edge-aware GAN EdgeConnect performs better than MISF, as its edge guidance helps preserve high-frequency details, whereas MISF produces blurrier restorations due to its dependence on a fixed filtering kernel. ShiftNet, on the other hand, tends to propagate artifacts because of feature shifting between layers. The diffusion-based ClarityNet has lower performance compared to GAN-based method DeepFillV2, as the former was trained for image synthesis rather than explicit artifact inpainting. Overall, our approach **performs better** than all the baselines despite being trained in an **unsupervised** manner on healthy images, primarily due to the diffusion auto-encoding setup, where the additional encoder effectively captures **contextual** information for artifact-free reconstruction.

*Vessel Segmentation on Synthetic Set*: To evaluate how well the restored images preserve vascular structures, we performed vessel segmentation using AutoMorph [18], specifically designed for fundus image analysis. The Dice scores for all the models are presented in the third column of Table 1. Our model achieves a higher Dice score, demonstrating superior vessel restoration. The better-performing baseline, DeepFillV2, achieves a relatively lower score due to artifact propagation near vessel boundaries. Among other baselines, EdgeConnect performs better than MISF owing to its edge awareness, while MISF produces blurrier vessel boundaries. ShiftNet tends to introduce unwanted structures, and ClarityNet generates partially occluded vessels, leading to degraded segmentation. Overall, our model **restores vascular structures** more effectively, as the iterative denoising-based

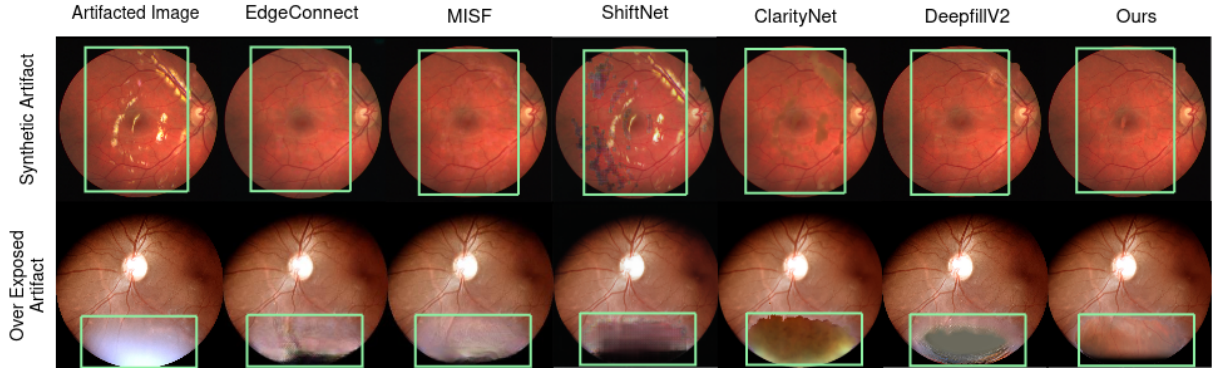

**Fig. 2**. Left to right: Comparison of restored images from various models. Top to bottom: Synthetic and Original artifact restorations. Green boxes highlight the region of artifact.

inpainting and context encoder jointly enhance vessel continuity and structural consistency.

*Quality Assessment on Test Set*: To evaluate restoration quality on real-world artifacts from mobile handheld acquisitions (without paired ground truth), we used the three-level AutoMorph grading system (*Good*, *Usable*, and *Reject*). The corresponding results are shown in the right segment of Table 1. Our model yields a **higher proportion** of *Good* images compared to all baselines. While the GAN-based DeepFillV2 and the diffusion-based ClarityNet perform relatively well, their scores remain lower than ours, as their generations, though visually realistic, often lack contextual fidelity. The remaining baselines (EdgeConnect, MISF, and ShiftNet) show lower performance due to their reduced effectiveness in restoring large artifacts, primarily present in the *Test Set* as over- and under-exposures.

*DR Classification on Restored Images*: The restored images were classified into diabetic retinopathy (DR) stages using an EfficientNet-B4 model trained on the *Train Set* combined with diseased samples from EyePACS. The results, reported in the last column of Table 1, show that our model achieves the highest classification accuracy. The baseline accuracy without restoration was $77.11\%$, indicating that images restored by our model enhance classification performance. Although EdgeConnect, ShiftNet, and DeepFillV2 show improved accuracy after restoration, their performance remains lower than ours, suggesting that our method **better** preserves contextual information necessary for **DR staging**.

*Ablation*: To analyze the role of the context encoder, we performed restoration under two settings: one using the representation ($z_1$) from the artifact image, and another using an interpolated representation $z_{\text{interp}} = 0.5z_1 + 0.5z_2$, where $z_2$ corresponds to the clean counterpart of the artifact image. The PSNR achieved using only $z_1$ is 38.98 dB, while using $z_{\text{interp}}$ yields 41.46 dB. This demonstrates that improved contextual **encoding** plays a **significant** role in enhancing the restoration quality.

### 4.2. Qualitative Analysis

The qualitative comparison of restoration performance across different methods is depicted in Figure 2. For synthetic artifacts, our model produces higher-quality restorations than all baselines. The GAN-based approaches, DeepFillV2 and EdgeConnect, replace vascular regions in the top-right area with yellowish patches, while MISF yields overall blurry restorations. The CNN-based ShiftNet and diffusion-based ClarityNet introduce undesired structures within the image. For overexposure artifacts, all models generate unnatural patterns, whereas our method achieves a more natural fundus-like appearance. These observations indicate that our model captures **contextual** information more **effectively** and is capable of handling artifacts of varying sizes and structures.

### 5. CONCLUSION

We have presented an unsupervised diffusion model for artifact restoration in mobile-acquired fundus images using a diffusion auto-encoding formulation that learns to generate high-quality images and employs inference-time inpainting for restoration. Quantitative evaluations demonstrate superior image quality and stronger contextual preservation compared to baseline methods. The restored images further enhance downstream analysis and disease diagnosis performance. Ablation studies validate that the encoded representations within the diffusion auto-encoder effectively guide the restoration process. While the current work employs a basic processing-based artifact mask generation pipeline, future extensions can integrate AI-driven artifact localization. Overall, our work demonstrates that diffusion auto-encoding-based formation can drive unsupervised restoration tasks in efficient manner.

### 6. COMPLIANCE WITH ETHICAL STANDARDS

The datasets used in this study are publicly available and anonymized. As no new data involving human subjects were collected, ethical review and informed consent requirements were waived.

## 7. CONFLICT OF INTEREST

The authors declare that they have no conflict of interest.